\documentclass[journal]{IEEEtran}

\usepackage{graphicx}

\usepackage{graphicx}
\usepackage{amsmath}
\usepackage{amssymb}
\usepackage{booktabs}
\usepackage{algorithm,algpseudocode}
\usepackage{soul}
\usepackage{xcolor}

\usepackage{multirow} 

\usepackage{hyperref}

\hypersetup{
    colorlinks=true,
    linkcolor=magenta,
    filecolor=magenta,      
    urlcolor=magenta,
    pdftitle={Overleaf Example},
    pdfpagemode=FullScreen,
    }

\ifCLASSINFOpdf
\else
\fi
\begin{document}
%
\title{Designed Dithering Sign Activation \\for Binary Neural Networks}
%
%
%

\author{Brayan~Monroy$^{*}$,
        Juan~Estupiñan$^{*}$,
        Tatiana~Gelvez-Barrera,
        Jorge~Bacca,
        and~Henry~Arguello
\\Department of Computer Science, Universidad Industrial de Santander\\ Bucaramanga, 680002, Colombia
\thanks{$^*$These are co-first authors with equal contributions. }

\texttt{
\url{ https://github.com/bemc22/DeSign  }
} \vspace{-1em}



}

%
%

\markboth{}%
{Shell \MakeLowercase{\textit{et al.}}: Bare Demo of IEEEtran.cls for IEEE Journals}
%



\maketitle

\begin{abstract}
Binary Neural Networks emerged as a cost-effective and energy-efficient solution for computer vision tasks by binarizing either network weights or activations. However, common binary activations, such as the Sign activation function, abruptly binarize the values with a single threshold, losing fine-grained details in the feature outputs. This work proposes an activation that applies multiple thresholds following dithering principles, shifting the Sign activation function for each pixel according to a spatially periodic threshold kernel. Unlike literature methods, the shifting is defined jointly for a set of adjacent pixels, taking advantage of spatial correlations. Experiments over the classification task demonstrate the effectiveness of the designed dithering Sign activation function as an alternative activation for binary neural networks, without increasing the computational cost. Further, DeSign balances the preservation of details with the efficiency of binary operations. 
\end{abstract}


%
\IEEEpeerreviewmaketitle

\section{Introduction}

Deep Neural Networks (DNNs) connote the state-of-the-art for most computer vision tasks, such as detection~\cite{szegedy2013deep}, classification~\cite{rawat2017deep}, or segmentation~\cite{guo2018review}. DNNs usually operate over hundreds to millions of real-valued ($32$-bit or $16$-bit) parameters, demanding expensive computational and storage resources~\cite{mcdanel2017embedded}. Binary neural networks (BNNs) connote an alternative that applies binarization strategies over the architecture parameters, including
weights~\cite{courbariaux2015binaryconnect}, activations~\cite{Kim2020BinaryDuo:}, or both~\cite{courbariaux2016binarized} to handle the complexity. 

A BNN employs binary values ($1$-bit) to perform most arithmetic operations as logical operations, significantly accelerating the running time (up to 52$\times$) and compressing the memory usage (up to 32$\times$) compared to DNNs. Nonetheless, the binarization produces a 
loss of precision because of the quantization error inherent in using binary values~\cite{xu2021recu} and the gradient mismatch of using binary-valued activations~\cite{Kim2020BinaryDuo:}.

Recent works have analyzed and adapted the statistical behavior of weights and activations to use binary weights while bridging the performance gap between BNNs and DNNs. For instance,~\cite{xu2021recu} introduces the Rectified Clamp Unit (ReCu) to solve the dead weights problem, where weights that reach a stationary state are not updated. Further, the real-valued Rectified Linear Unit function (ReLU) has been approximated using strategies based on the binary-valued Sign function~\cite{montesinos2022fundamentals}. To name,~\cite{rastegari2016xnor, lin2020rotated} modify the forward and backward steps of the Sign activation by introducing learned non-binary parameters,~\cite{cai2017deep} quantizes the ReLU to a fixed amount of discrete levels, and~\cite{zhang2022dynamic, liu2020reactnet} learn the shifting in the ReLU and Sign activations to minimize the bits of representation and preserve the precision.

Besides affecting the performance, the loss of precision causes the loss of fine-grained details in the output. This issue is a recurrent phenomenon in binary image representation, mitigated through dithering strategies that adjust the density of binary values in the output image to closely approximate the average gray-level distribution of the original image~\cite{schuchman1964dither, holesovsky2017compact}. Dithering introduces controlled variations in the form of noise or designed patterns following different schemes. Fixed dithering compares all pixels to a uniform threshold, ensuring consistent thresholding across the entire image~\cite{ulichney1987digital}. Random dithering compares the pixels to randomly generated thresholds, avoiding homogeneous areas~\cite{6773262}. Iterative dithering utilizes error diffusion and applies to dither iteratively over pixels and their neighbors~\cite{ulichney1999review}. Ordered dithering utilizes fixed dither matrices, such as the halftone or Bayer, taking advantage of spatial information to produce specific dithering effects~\cite{ulichney1999void}.

\begin{figure}[!tb]
 \centering
 \includegraphics[width=\linewidth]{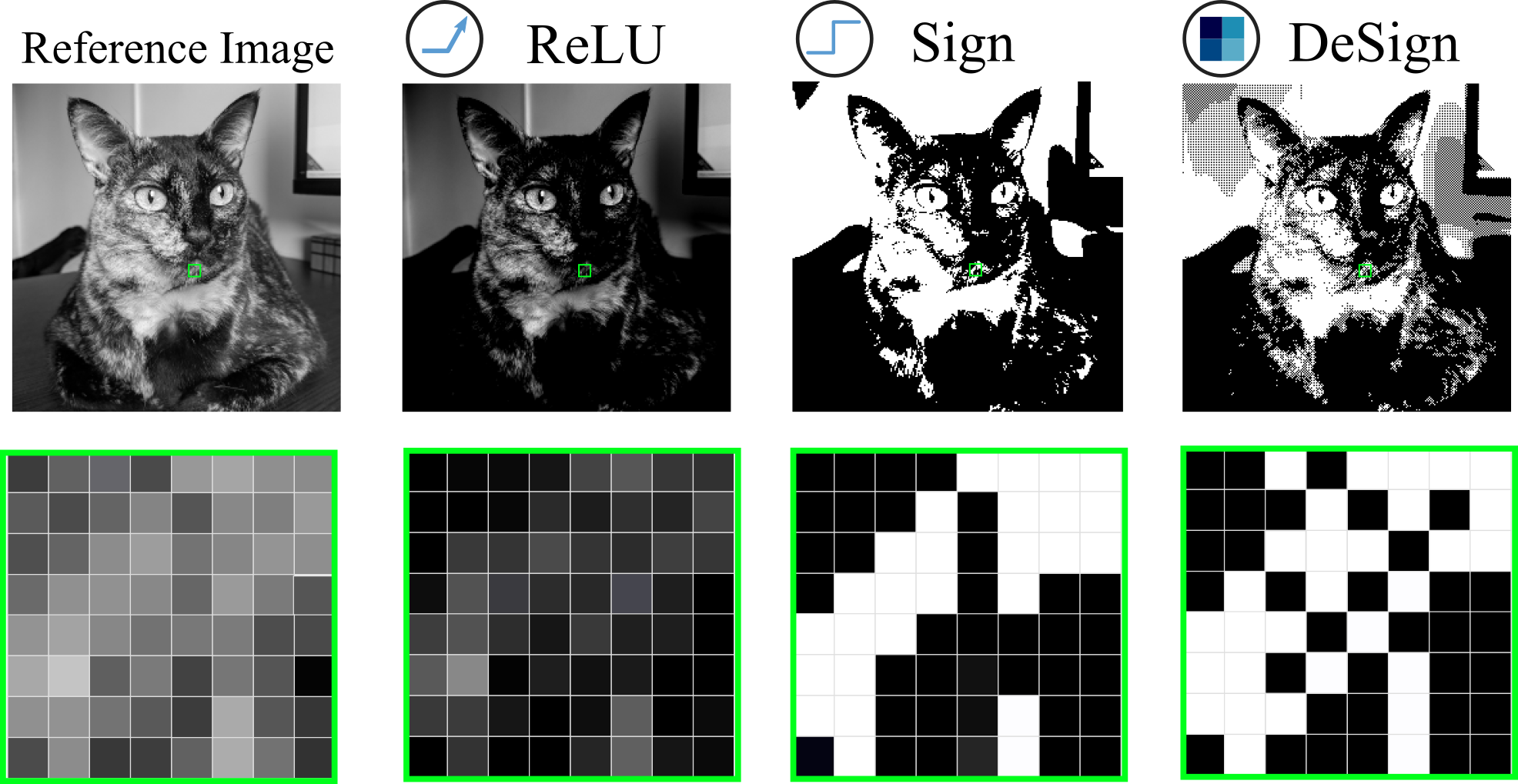}
 \caption{Illustration of the output when applying the ReLU, Sign, and proposed DeSign activations to a reference image. \textit{(Top)} Generated activation maps. \textit{(Bottom)} Zoom of a specific output patch. Although Sign and Design outputs are entirely binary, Design offers a better representation of the structure and preservation of fine-grained details within the image.} 
 \label{fig:reluapro}
\end{figure}

Inspired by the approximation of real-valued activations using the Sign function and the dithering process in binary image representation, this work proposes a designed dithering Sign (DeSign) activation for BNNs. DeSign incorporates periodic spatial shifts to encode the most relevant information from binary convolutions. Unlike previous methods that quantize the ReLU to low-bit precision levels or learn shifting parameters along the features independently, DeSign employs a threshold kernel whose values are designed jointly for a set of adjacent pixels taking advantage of local spatial correlations. Then, the kernel is repeated periodically across the spatial dimension. The design methodology comprises two steps (\romannumeral 1) the optimal threshold kernel is selected as the one that maximizes an objective function that measures the ability to preserve structural information. (\romannumeral 2) the entries are re-scaled to match the distribution of operations in the BNN forward propagation.

Figure~\ref{fig:reluapro} illustrates the effect of applying the ReLU, Sign, and DeSign activations to a real-valued reference image. As expected, the ReLU preserves most of the structure by retaining real values. Conversely, Sign activation diminishes most structural details by mapping values to only two levels. DeSign lies between ReLU and Sign, with superior preservation of structural details compared to Sign while mapping only to two levels, as shown in the zoomed window.

Simulations over CIFAR-10, CIFAR-100, and STL-10 classification datasets and two state-of-the-art BNN architectures validate the effectiveness of DeSign to boost overall BNN accuracy while preserving binary operations without adding significant computational load. DeSign also mitigates the influence of real-valued learned layers, such as batch normalization, enhancing baseline BNNs accuracy in up to $4.51\%$. The main contributions of this work can be summarized as follows \begin{enumerate}
\item DeSign, a binary-designed dithering activation based on a spatially periodic threshold kernel that shifts the Sign activation to preserve the structure and fine-grained details, described in Section~\ref{sec:proposedDeSign}.
\item An optimization-based methodology to design the threshold kernel taking advantage of spatial correlations described in Section~\ref{sec:pattern}.
\item A performance improvement for classification task compared to literature binary approaches and a drawing up to real-valued networks performance, described in Section~\ref{sec:simulations}.
\end{enumerate}


\section{Binary Neural Networks Background}
\label{sec:bnnbackground}
The BNN forward propagation involves three layers: (\romannumeral 1) A binary convolution layer convolving a binary input with a binary kernel, (\romannumeral 2) a batch normalization layer that standardizes the information range, and (\romannumeral 3) an activation layer that maps the output values to a different space. The activation can be real-valued to mitigate the loss of information or binary-valued to obtain a full BNN.

\subsection{Binary Convolution Layer} 
\label{sub:binary_conv}
The binary convolution layer performs bit-wise operations acting as the logic gate XNOR~\cite{rastegari2016xnor}. Let $\mathbf{K} \in \mathbb{Z}_2^{k \times k}$ denote the binary kernel of size $k \times k$, $\mathbf{X} \in \mathbb{Z}_2^{h \times w}$ denote a binary matrix of size $h \times w$, and $\mathbb{Z}_2$ denote a binary set with two elements. In this paper, the binary set is selected as $\mathbb{Z}_2 = \{-1,1\}$. Then, the binary convolution is given by \begin{equation}
\begin{split}
\mathbf{X}_c= \mathbf{K} \circledast \mathbf{X},
\end{split}
\label{eq:binaryconvolution}
\end{equation} where $\circledast$ denotes the 2D convolution operator and $\mathbf{X}_c \in \mathbb{Z}^{h-k+1 \times w-k+1}$ denotes the convolved output. In this manner, $\circledast: \mathbb{Z}_2 \rightarrow \mathbb{Z}$ denote the convolution operator convolving a binary kernel with a binary matrix. The range of this operator depends on the kernel size $k$ and can be defined as 
\begin{equation}
\mbox{range}(\circledast) = \left\{ i \in \mathbb{Z} \; \middle| \; i = -k^2 + 2\ell, \, \forall \ell \in \mathbb{Z} \cap [0, k^2] \right\}.
\end{equation}
For instance, for $k = 3$ the entry values of $\mathbf{X}_c$ will belong to the set $\{ -9 + 2\ell : \ell \in \mathbb{Z} \cap [0 \hspace{2mm} 9]\}$.

For ease of notation, this forward process is presented as a 2D process, however, it should be repeated across all features of the input matrix.

\subsection{Batch-normalization Layer} 
The convolution operation in the binary layer produces integer values even when the weights and activations are binary. Hence, the batch-normalization layer is used between the convolution and activation layers to prevent unstable data flowing~\cite{ioffe2015batch}. The batch normalization standardizes the output of the previous layer as follows 
\begin{equation}
 \mathbf{X}_s = \frac{\mathbf{X}_c - \mu}{ \sqrt{\sigma + \epsilon} }\cdot \gamma + \beta.
 \label{eq:batch}
\end{equation} 
In \eqref{eq:batch} the convolved image $\mathbf{X}_c$ is standardized to a normal distribution $\mathcal{N}(0, 1)$ given the moving average and variance estimation $\mu, \sigma$. Then, the trainable parameters $\gamma, \beta $ re-scale the distribution of the normalized input to a Gaussian distribution, such that $\mathbf{X}_s \sim \mathcal{N}(\beta, \gamma)$. Remark that there is one different $\gamma$ and $\beta$ operating along the features of the input $\mathbf{X}_c$, namely each feature of the normalized input is re-scaled to a different mean and deviation Gaussian distribution.

\subsection{Activation Layer}
BNNs can employ Real-valued or Binary-valued activations. The Real-valued activation offers superior performance but forfeits the ability to harness 1-bit operations inherent to BNNs. Conversely, Binary-valued activation preserves the utilization of 1-bit operations but substantially decreases the performance. Subsequent subsections explore the most used activations within the context of BNNs.

\begin{figure*}[t!]
\centering
\includegraphics[width=\linewidth]{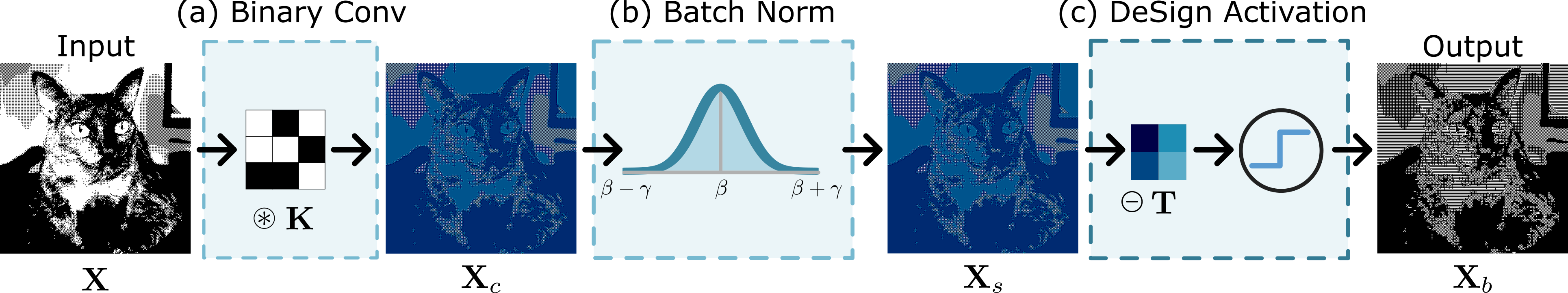} 
\caption{
Binary forward propagation scheme with the proposed DeSign activation. (a) The input $\mathbf{X} \in \mathbb{Z}_2^{h \times w}$ is convolved with binary kernels $\mathbf{K} \in \mathbb{Z}_2^{k \times k}$. (b) The output $\mathbf{X}_c$ is batch-normalized using the trainable parameters $\gamma$ and $\beta$ through the features. (c) The batch-normalized output $\mathbf{X}_s$ is passed trough the DeSign activation. Precisely, the threshold kernel $\mathbf{T}$ is incorporated in the third layer, through the operation $\mathbf{X}_s-(\mathbf{T} \otimes \textbf{1}$) to impose a dithering structure that helps in the preservation of information. Then, the conventional Sign activation is applied, obtaining the binary output $\mathbf{X}_b \in \mathbb{Z}_2^{h-k+1 \times w-k+1}$.}
\label{fig:design2}
\end{figure*}

\subsubsection{Real-valued Activation} 
\label{sec:relu}
The Rectified Linear Unit $\text{ReLU}(\cdot): \mathbb{R} \rightarrow \mathbb{R}_{0}^{+} $ is a real-valued activation that maintains equal positive values and transforms negative ones to zero as follows
\begin{equation}
\label{eqn:relu}
\text{ReLU}(x)= \text{max}(0, x).
\end{equation}

The range of the ReLU corresponds to all non-negative real-values so that the range of applying the ReLU at the output of the convolution operator $\circledast$ in \eqref{eq:binaryconvolution} can be expressed as
\begin{equation}
\Omega = \left\{ i \in \mathbb{Z}_0^{+} \; \middle| \; i = -k^2 + 2\ell, \, \forall \ell \in \mathbb{Z} \cap \left[ \frac{k^2}{2} \hspace{2mm} k^2 \right] \right\}.
 \label{eqn:Omega_set}
\end{equation}

\subsubsection{Binary-valued Activation}
The binary-valued activation is habitually divided into one function for the forward pass and one for the backward pass to avoid quantization drawbacks such as discontinuities or vanishing gradients. The Sign activation $\text{Sign}(\cdot): \mathbb{R} \rightarrow \mathbb{Z}_2$ is a piece-wise function commonly used in the forward pass that returns the sign of the input given by
\begin{equation}
\label{eqn:sign}
\text{Sign}(x)= \frac{x}{\vert x \vert}.
\end{equation} 
The constant behavior of the Sign function in \eqref{eqn:sign} produces a zero derivative throughout its domain. Then, the Clip function defined in \eqref{eqn:clip} is used during the backward pass to approximate the Sign derivative, ensuring the network parameters update~\cite{qin2020binary}.
\begin{equation}
\label{eqn:clip}
 \text{Clip}( x ) = \text{max}( -1 , \text{min}(1, x) ).
\end{equation}


\section{DeSign: Designed Dithering Sign Activation}
\label{sec:proposedDeSign}
Inspired by the dithering process that adjusts the density of binary values to approximate the average gray-level distribution, this paper proposes the DeSign activation with periodic spatial shifts for BNNs. DeSign aims to reduce the information loss while maintaining 1-bit operations. Mathematically, $\text{DeSign}(\cdot): \mathbb{R} \rightarrow \mathbb{Z}_2$ is defined as 

\begin{equation}
\mathbf{X}_b = \text{DeSign}(\mathbf{X}_s; \mathbf{T})=\text{Sign}(\mathbf{X}_s-(\mathbf{T} \otimes \textbf{1})),
\label{eqn:design} 
\end{equation}
where $\mathbf{T} \in \mathbf{\Omega}^{d \times d}$ $d > 1$ is a threshold kernel, shifting the Sign activation for each pixel in the input before the binarization, $\otimes$ denotes the Kronecker product and $\textbf{1}$ denotes a one matrix fitting the input dimension $\mathbf{X}_s$. Note that using a threshold kernel acting for a small spatial window of size $d \times d$ enables leveraging inherent local spatial correlations in comparison to using an independent threshold for each location. In addition, the domain of the threshold kernel is constrained to the set $\Omega$, based on the definition in \eqref{eqn:Omega_set}, provided that this set contains all possible quantization levels.

The structure of $\mathbf{T}$ determines the performance of the BNN, so that we propose to design $\textbf{T}$ in such a manner that it preserves as much as possible the distribution of the values produced by binary convolutions as if the ReLU were applied. The proposed design taking advantage of spatial correlations, is presented in Section~\ref{sec:pattern}. 

Remark that the proposed DeSign activation can be incorporated in the forward propagation of any BNN, as schematized in Fig.~\ref{fig:design2}, where the spatially periodic threshold kernel $\mathbf{T}$ is included between the batch normalization layer and the conventional Sign activation layer.

\section{Threshold Kernel Design}
\label{sec:pattern}
The threshold kernel $\mathbf{T}$ can be defined in one of several ways: it can be established ad-hoc, drawn from existing literature patterns~\cite{holesovsky2017compact}, or crafted through an optimization approach, such as the methodology outlined in this paper. In practice, the design methodology encompasses two steps. (\romannumeral 1) the selection of the optimal thresholding kernel $\mathbf{T}$ maximizing an objective function that quantifies the preservation of structural information detailed in Section \ref{subsec:kernel_pattern}, and (\romannumeral 2) the scaling of the threshold entries to align with the batch-normalization process described in Section \ref{subsec:kernel_values}.

\subsection{Threshold Kernel Selection}
\label{subsec:kernel_pattern}
The proposed optimization methodology aims to select a threshold kernel that preserves structural information based on the behavior of BNNs forward propagation analyzed in Section \ref{sec:bnnbackground}, in particular, in view of the range obtained when using the ReLU given by the set $\Omega$ in \eqref{eqn:Omega_set}.

For preserving structural information, DeSign should be able to maintain the largest possible difference between adjacent pixels within the kernel window across all assessed spatial windows encompassing the input image. Consequently, we propose maximizing the expected total variation in the output resulting from applying the ReLU to the DeSign. 

\begin{figure*}[!t]
 \centering
 \includegraphics[width=\linewidth]{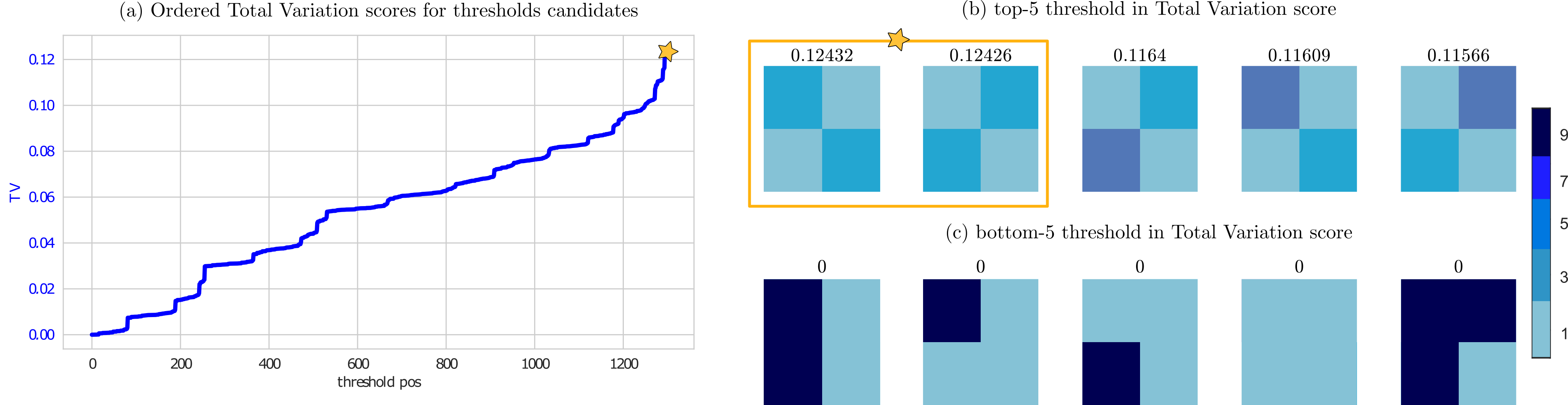}
 \caption{
 Total Variation score of all threshold kernel candidates. (a) Ordered TV score, (b) top-5 threshold kernels with the highest TV score, and (c) bottom-5 threshold kernels with the lowest TV score. }
 \label{fig:threseval}
\end{figure*}

Mathematically, we express the optimization problem to find the optimal thresholding kernel $\mathbf{T}^* \in \mathbf{\Omega}^{d \times d}$ as follows

\begin{equation}
\textbf{T}^* \in \underset{\textbf{T} \; \in \; \mathbf{\Omega}^{d \times d} }{\text{arg max}} \quad \mathbb{E}_{\textbf{X}}\Big[  \Big\Vert \text{ReLU}(\underbrace{\text{Sign}(\mathbf{X} \circledast \mathbf{K} - (\textbf{T}\otimes \mathbf{1}))}_{\text{DeSign}(\mathbf{X};\mathbf{T})}) \; \Big\Vert  _{\text{TV}} \Big]. 
\label{eq:optimization_fn}
\end{equation} $||\cdot||_{\text{TV}}$ denotes the Total-Variation (TV) operator~\cite{gelvez2020nonlocal}.

This paper employs a brute force strategy for solving~\eqref{eq:optimization_fn} since the involved variables are all discrete and the domain of the optimization variable $\mathbf{T}$ is constrained to the set $\Omega$ with cardinality $|\Omega| = \lceil {k^2}/{2}^{}  \rceil$. Let $\mathcal{T}$ be the set containing all possible threshold kernels of size $d \times d$ whose entries can take the values in the set $\Omega$; the cardinality of $\mathcal{T}$ is given by $|\mathcal{T}| = {\lceil {k^2}/{2}^{}  \rceil}^{d^2}$, i.e., it is reasonable to evaluate all possible thresholds kernels for small values of $k$ and $d$.

The computation of the TV score for each particular threshold $\mathbf{T} \in \mathcal{T}$ is done by simulating multiple binary convolutions as described in section \ref{sub:binary_conv}, using random binary kernels $\mathbf{K}$ for all images $\mathbf{X}$ that belong to a given dataset.

Once all TV scores have been computed for each kernel, the candidates are arranged in ascending order, based on their respective TV scores. The quantitative findings for the case of $k=3$, $\Omega = \left\{0, 1, 3, 5, 7, 9 \right\}$,  $d = 2$, $|\mathcal{T}| = 1296$, and using the CIFAR-100 dataset are presented in Figure~\ref{fig:threseval}(a). To facilitate the visual analysis, we focus on the top five kernel candidates with the highest TV value, as depicted in Figure~\ref{fig:threseval}(b). These top five kernels exhibit a Bayer filter spatial distribution on the $1$, $3$, and $5$ levels. The first and second kernels, which possess the highest TV scores, represent permutations of the $\mathbf{T}^*=[1 ,1 ,3 ,3]$ tile configuration. Therefore, we select such kernel as the optimal one maximizing the expected TV according to our methodology to be used in the subsequent simulations.

\subsection{Entry Scaling to Batch Normalization}
\label{subsec:kernel_values}

\begin{figure}[!t]
 \centering
 \includegraphics[width = \linewidth]{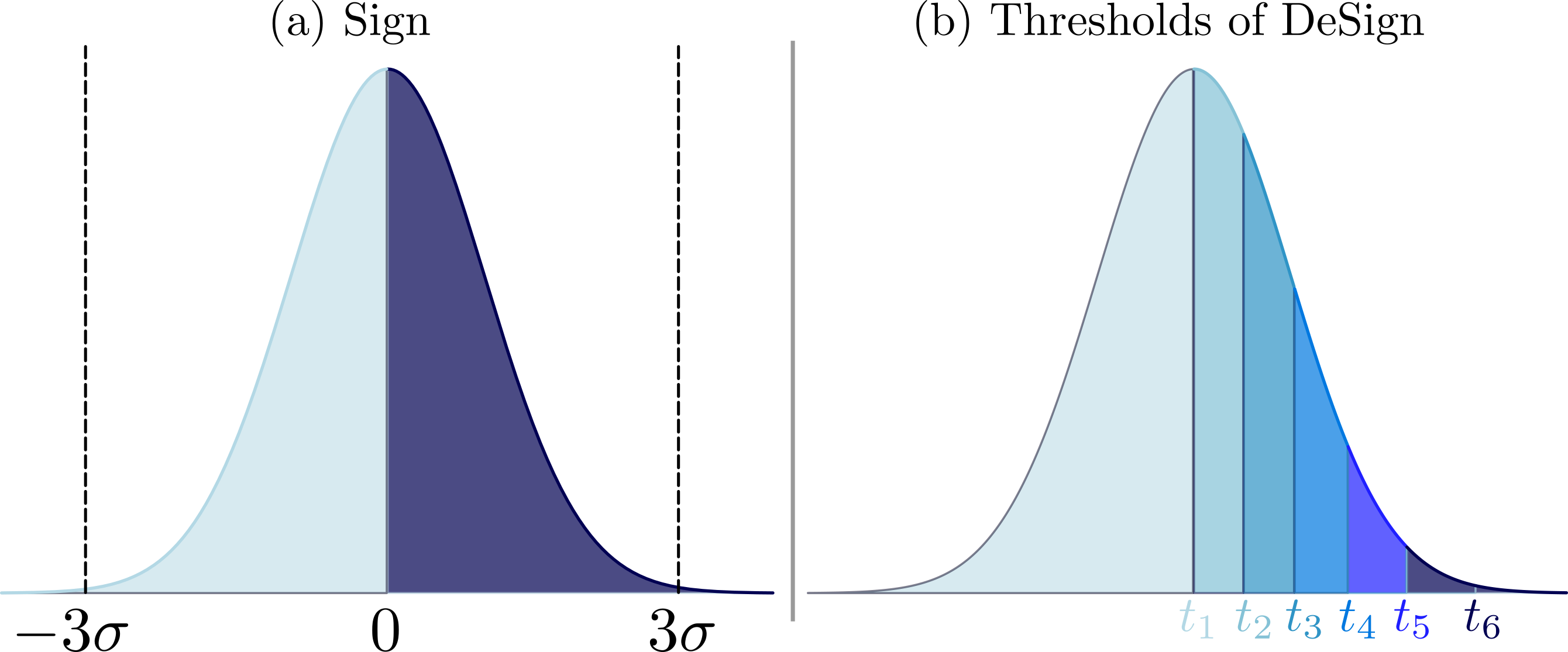} \vspace{-1em}
 \caption{
 Distribution range estimation: 1) When using the Sign, there are only three options, all numbers negative, i.e., $[-3\sigma, 0]$, all positive i.e., $[0, 3\sigma ]$, or combined i.e., $[-3\sigma , 3\sigma]$. 2) When using the proposed thresholds, different ranges are possible using the reference threshold values $t_\kappa$, increasing the precision and approximating the behavior of the ReLU function.}
 \label{fig:distribution_range} 
\end{figure}

Including batch normalization layers into BNNs induces the mapping of integer outputs from binary convolutions to floating-point values. Hence, for implementation purposes, the entries of the selected threshold kernel $\mathbf{T}^*$ have to be adapted to align with the normalized data distribution. This paper adopts the methodology outlined in~\cite{cai2017deep}, which utilizes the Half-wave Gaussian quantization technique. Specifically, the right side of the Gaussian distribution corresponds to the behavior associated with ReLU activations is used to select $N = |\Omega| =  \lfloor k^2/{2}^{}  \rfloor + 1$ quantization levels, matching the $N$ levels generated by the binary convolution (Section~\ref{sec:relu}). Subsequently, a K-means algorithm is employed to quantize the Half-wave Gaussian distribution into $N$ clusters as illustrated in Figure~\ref{fig:distribution_range}(b). Finally, the process of mapping the quantization levels to their corresponding real values involves replacing the quantization levels obtained in the threshold pattern $\textbf{T}$ with their left-side threshold values. These threshold values are determined based on the results obtained through the K-means algorithm and define the scaled real-valued~~threshold $\textbf{T}_s$.

Figure~\ref{fig:distribution_range} compares the distribution range for the Sign and DeSign scenarios, illustrating how using DeSign improves the preservation of structural information in comparison to Sign. Precisely, the Sign distribution only has one quantization level so that it can only discriminate between two fixed groups. In contrast, the Design distribution increases the possible ranges by selecting different threshold values.

\begin{figure*}
    \centering
    \includegraphics[width=\linewidth]{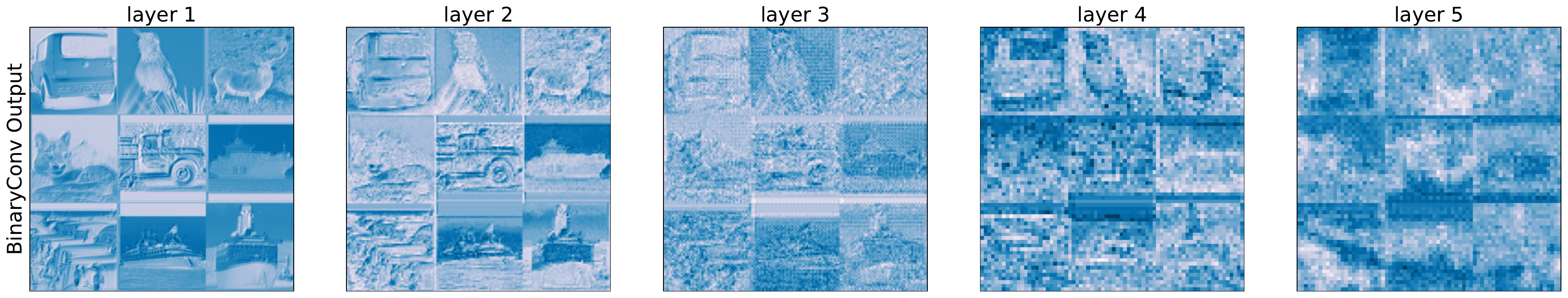}
    \includegraphics[width=\linewidth]{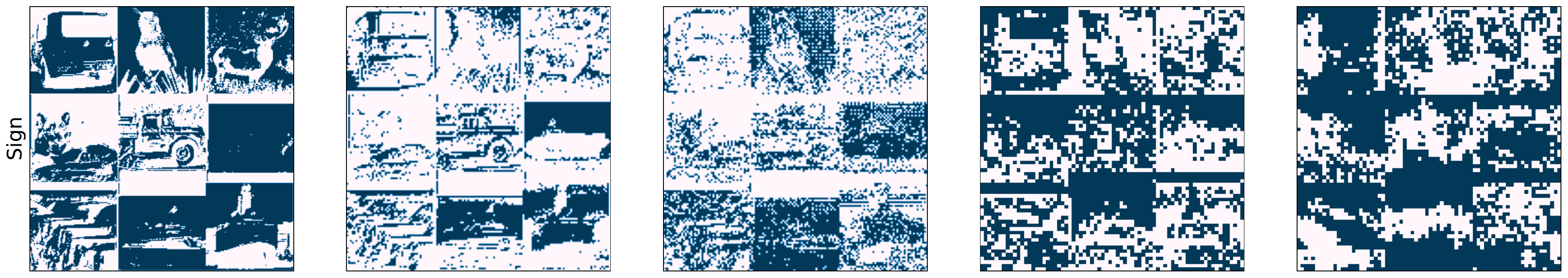}
    \includegraphics[width=\linewidth]{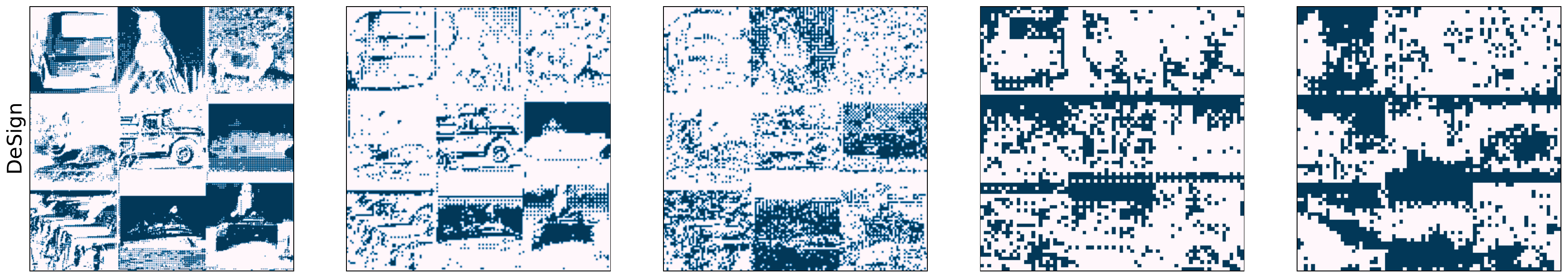}
    \caption{Middle outputs activations of BNN architecture on STL-10 dataset. (\textit{Top}) Binary Convolution outputs, (\textit{Middle}) Sign activations outputs, and (\textit{Bottom}) DeSign activations outputs. The incorporation of DeSign activations enables the preservation of fine details along BNNs whiout additional computational cost.}
    \label{fig:middleouts}
\end{figure*}

One further remark on how DeSign preserves the fine-grained details across the middle layers in the architecture is illustrated in~Fig.~\ref{fig:middleouts}. It can be observed that in comparison to Sign, the binary representation provided by the DeSign activation keeps the structure of the binary convolution output. Notice for instance in the layer 1 how the DeSign activation can still preserve the structure of elements that are completely destroyed with the Sign activation such as the ship.

\subsection{3D scenario Design}
\label{sec:designdesign3D}
The proposed 2D threshold kernel design in Section~\ref{sec:pattern} can be extended to take advantage of inter-channel correlations and promote channel variety when dealing with images where each channel represents a different modality, such as color, texture, or class. Thus, instead of using the same threshold kernel to each channel, we present a technique that applies a different threshold kernel to each channel, referred to as DeSign3D, increasing the network's ability to capture unique characteristics and diverse information per channel. DeSign3D is built upon the foundation of the 2D design, where we utilize two techniques based on the set $\Omega$ to generate a 3D threshold as follows.

\begin{itemize}
 \item \textbf{Circular shift:} This technique employs a 2D threshold and shifts the values of the set $\Omega$ to generate an additional channel, i.e., each channel is created as a circularly shifted version of the previous channel.
 \item \textbf{Complement:} This technique assigns complementary thresholds to generate an additional channel, where the complement of the threshold $t_{\kappa}$ given by $\Omega[\kappa]$, is calculated by $\Omega[{(N-[\kappa-1])}]$.
 \end{itemize}

\section{Simulations and Results}
\label{sec:simulations}
This section outlines the experiments conducted to validate the efficacy of the DeSign activation. First, Section~\ref{sec:design3} presents a comprehensive benchmark analysis of state-of-the-art BNNs on several classification datasets, comparing the performance of the 2D and 3D variants of the DeSign activation. The influence of batch normalization over the DeSign activation is also analyzed. Finally, a state-of-the-art comparison using various binarization strategies demonstrates how incorporating the DeSign activation enhances the overall performance. The experiments aim to demonstrate the effectiveness of the DeSign activation in improving the accuracy of BNNs, through a rigorous comparison methodology to ensure that the presented results are accurate and reliable, contributing to the development of more efficient and accurate BNNs.

\subsection{Comparison Benchmark}
\label{sec:design3}

The benchmark consists of evaluating the influence of using the proposed DeSign between the batch normalization and activation layers. For this, the VGGsmall and ResNet18 architectures were used across the BNN~\cite{courbariaux2016binarized}, and ReCU~\cite{xu2021recu} binarization strategies, on the classification task in the CIFAR-10, CIFAR-100, and STL-10 datasets. The performance is evaluated using the accuracy metric.

\subsection{Selection of Design Strategy}

Multiple experiments were carried out to determine the kernel size and the possible design scenarios, i.e., 2D and 3D.
The experimental setup consists of 3 runs of the BNN~\cite{courbariaux2016binarized} method for each binary activation case in the CIFAR-10 dataset on fixed and learnable batch-normalization parameters. The evaluated cases for DeSign can be grouped into three categories summarized as follows
\begin{enumerate}
 \item DeSign 2D: The design threshold kernel is broadcasted along the dimension of the features.
 \item DeSign 3D: The design threshold kernel is permuted and reflected to generate 3D threshold patterns. In this method, C stands for complementary thresholds, and S for the circular shift on the designed threshold kernel.
 \item Learned 2D: The threshold kernel is learned as a parameter of the network using the end-to-end scheme presented in~\cite{bacca2021deep}.
\end{enumerate}

 \begin{table}[!t]
 \centering 
 \normalsize
 \caption{Comparison of different DeSign 2D and 3D strategies on CIFAR-10 datasets with the BNN~\cite{courbariaux2016binarized} method. Kernel size $k=\{2, 3\}$ on fixed $\text{BN}(0/1)$ and learned $\text{BN}(\beta/\gamma)$ batch-normalization settings.}
 \resizebox{\columnwidth}{!}{
 \begin{tabular}{l|c|c c| c c }
 \hline
 \multirow{2}{*}{\textbf{Dither Method}} & \multirow{2}{*}{k}
 & \multicolumn{2}{c|}{ $0 /1 $ } 
 & \multicolumn{2}{c}{ $\beta / \gamma$ } \\ 
& & Last & Best & Last & Best \\ \hline \hline
w/o Dither~\cite{courbariaux2016binarized}& - &	85.83	& 85.97	& 90.41	& 90.70 \\
Random 2D & 2 &	85.72	& 85.95	& 89.33	& 89.68 \\
Learned 2D & 2 &	89.92	& 90.23	& 90.46	& 90.75 \\
DeSign 2D & 2	& 89.70	& 89.75	& 89.82	& 90.09 \\
DeSign 3D-C	& 2 & 90.30	& 90.46	& 90.58	& 90.98 \\
DeSign 3D-S & 2	& \textbf{90.36}	& \textbf{90.48}	& 90.84	& \textbf{91.09} \\
DeSign 2D & 3	& 89.63	& 89.89	& 90.81	& 90.83 \\
DeSign 3D-C	& 3 & 89.97	& 90.29	& \textbf{90.97}	& 91.07 \\
DeSign 3D-S & 3	& 89.80	& 89.87	& 90.68	& 90.97 \\ \hline
 \end{tabular}} 
 \label{tab:design2d3d} 
\end{table}

\begin{table}[!t] \small \centering 
\caption{Batch normalization influence on \\ CIFAR-10, CIFAR-100 and STL-10 Datasets. \label{tab:batch10}}
 \resizebox{\columnwidth}{!}{
\begin{tabular}{l|c|cc|cc}
 \hline
\multirow{2}{*}{\textbf{Dataset} \hspace{-1em} } 
& \multirow{1}{*}{\textbf{Batch}} 
& \multicolumn{2}{c|}{\textbf{BNN\cite{courbariaux2016binarized} }} 
& \multicolumn{2}{c}{\textbf{ReCU~\cite{xu2021recu} }} \\
 & \textbf{Setting}
 & \textbf{Baseline} \hspace{-1em} 
 & \textbf{DeSign} 
 & \textbf{Baseline} \hspace{-1em} 
 & \textbf{DeSign}  
 \\ \hline  \hline
 \multirow{4}{*}{CIFAR-10}&
$ 0/1 $ & 
86.44 & \textbf{90.87} & 
89.81 & \textbf{91.02} 
\\
 &
 $\beta /1 $ &
91.15 & \textbf{91.34} & 
91.00 & \textbf{91.54} 
\\
 &
$0/\gamma$ &
86.51 & \textbf{90.22} &
89.87 & \textbf{92.06}
\\
 &
$\beta/\gamma$ &
90.88 & \textbf{91.06} & 
92.84 & \textbf{92.92} 
\\ \hline
 \multirow{4}{*}{CIFAR-100}&
$ 0/1 $ 
& 55.54 & \textbf{63.13} 
& 66.01 & \textbf{69.57} 
\\
 &
 $\beta /1 $
&62.62 & \textbf{64.19} 
& 69.03 & \textbf{69.77}  
\\
 &
$0/\gamma$ 
& 55.93 & \textbf{63.67} 
& 66.22 & \textbf{69.90}  
\\
 &
$\beta/\gamma$
& 62.23 & \textbf{64.30} 
& 72.36 & \textbf{72.47} 
\\ \hline
 \multirow{4}{*}{STL-10}&
$ 0/1 $  
& 68.51 & \textbf{73.09}
& 75.00 & \textbf{82.05} 
\\
 &
 $\beta /1 $
& 67.74 & \textbf{73.31}
& 78.40 & \textbf{82.55}  
\\
 &
$0/\gamma$ 
& 68.46 & \textbf{73.62}
& 75.45 & \textbf{84.12}
\\
 &
$\beta/\gamma$
& 68.64 & \textbf{73.24}
& 85.16 & \textbf{85.87}
\\ \hline 
\end{tabular} } 
\end{table}

Table~\ref{tab:design2d3d} presents the quantitative comparison of the evaluated scenarios, presenting the metrics obtained in the last and best epoch. The results show that incorporating a feature-level bias using DeSign 3D activation significantly improves the overall network performance compared to using only DeSign 2D. After conducting experiments on the CIFAR-10 dataset, we determined that the highest accuracy was achieved with a spatial kernel of size $k=2$ and a circular shift along the feature dimension to generate DeSign 3D. We selected the DeSign 3D-S spatial pattern for the subsequent state-of-the-art comparisons based on these results. Notably, DeSign 3D activation reduces the influence of batch-normalization parameters on network accuracy, achieving a base accuracy of $90.48\%$ compared to $85.97\%$ for the baseline BNN method that uses learned batch-normalization parameters, thus demonstrating the efficacy of DeSign 3D in mitigating the effects of learned batch normalization parameters, remarked in the similarity of the results in the column of fixed $\text{BN}(0/1)$ and learned $\text{BN}(\beta/\gamma)$ batch-normalization settings. Since DeSign does not increase the amount of learnable parameters, it offers a promising solution for improving BNNs performance without adding significant computational burden.

\subsection{Batch-normalization Influence Analysis}

This section examines the impact of learning the mean ($\beta$) and variance ($\gamma$) parameters of the batch-normalization layer on the performance of the proposed method. Specifically, the configuration using a fixed $\mathcal{N}(0,1)$ distribution with no learned parameters is compared against to a configuration that incorporates both learned parameters, resulting in a $\mathcal{N}(\beta,\gamma)$ distribution.

Table~\ref{tab:batch10} reports the obtained results, showing that normally the absence of learned parameters significantly impacts the network performance, i.e., using a distribution $\sim\mathcal{N}(0,1)$ leads to a drastic reduction in the performance compared to using both learned parameters. Contrary, when using the proposed DeSign, it can be observed a reduction in the performance gap between no-learned and learned batch-normalization configurations, for all evaluated methods and datasets. These results support the claim that DeSign preserves the spatial bias across the activations of batch-normalized neural networks, improving the performance.

In addition, the individual influence of learning the mean $\beta$ and deviation $\gamma$ of the batch normalization with DeSign is evaluated. As it can be seen in Table~\ref{tab:batch10}, the use of a learned means $\mathcal{N}(\beta, 1)$ increases the obtained performance even more, which can be interpreted as a combination of the spatial bias provided by DeSign and a feature bias provided by batch normalization.

For the case of learning the deviation $\gamma$, it can be noticed a performance drop in preliminary results. Analyzing the relationship between the designed thresholds of the DeSign activation and the influence of re-scaling the values of its input,  it becomes evident that, while the input of the DeSign function now corresponds to a distribution $\mathcal{N}(0, \gamma)$, the thresholds were designed to reduce the quantization error for a distribution $\mathcal{N}(0, 1)$. Thus it is necessary to re-scale the values of the thresholds to preserve the quantization intervals for which they were designed. Scale correction on threshold values was used for all the cases where the learning of the $\gamma$ deviation was employed.

Once the re-scaling was incorporated into the thresholds, there was no statistically significant increase in performance for BNN~\cite{courbariaux2016binarized} method, while for ReCU, there was a significant increase in performance in all datasets.

These findings suggest that synergy between spatial bias provided by DeSign activation and feature-level scale and bias induced by batch normalization could boost even more binarization methods that exploit information entropy and standardization of weights, such as~ReCU~\cite{xu2021recu}.

\begin{table}[!t] \normalsize \centering
\caption{Performance comparison with the state-of-the-art on CIFAR-10. W/A denotes the bit length of the weights
and activations. FP is short for full precision. \label{tab:sota}}
 \resizebox{\columnwidth}{!}{
\begin{tabular}{l|c|c|c} 
\hline
\textbf{Network} & \textbf{Method} & \textbf{W/A} & \textbf{Top-1} \\ \hline \hline
\multirow{6}{*}{ResNet-18} & FP & 32/32 & 94.8 \\
 & RAD~\cite{ding2019regularizing} & 1/1 & 90.5 \\
 & IR-Net~\cite{qin2020forward} & 1/1 & 91.5 \\
 & RBNN~\cite{lin2020rotated}& 1/1 & 92.2 \\
 & ReCU~\cite{xu2021recu} & 1/1 & 92.8 \\
 & \textbf{ReCU* (Proposed)} & 1/1 & \textbf{92.9} \\ \hline
\multirow{6}{*}{VGG-small} & FP & 32/32 & 94.1 \\
 & XNOR-Net~\cite{rastegari2016xnor} & 1/1 & 89.8 \\
 & DoReFa~\cite{zhou2016dorefa} & 1/1 & 90.2 \\
 & IR-Net~\cite{qin2020forward} & 1/1 & 90.4 \\
 & BNN~\cite{courbariaux2016binarized} & 1/1 & 90.9 \\
 & \textbf{BNN* (Proposed)} & 1/1 & \textbf{91.3} \\ \hline
\end{tabular}} 
\end{table}

\subsection{Classification Performance Comparison.} As DeSign aims to increase the information preserved along binarization neural networks, it could be introduced by substituting vanilla binarization functions such as Sign. 
The performance of DeSign is evaluated in comparison to vanilla binarization functions, such as Sign, by substituting them in the DNN~\cite{courbariaux2016binarized} and ReCU~\cite{xu2021recu} binarization methods, applied to VGG-small and ResNet-18 network architectures, respectively. The quantitative comparison, presented in Table~\ref{tab:sota} demonstrates a performance improvement while preserving the total network parameters and 1-bit operations.

It is worth noting that BNN prioritizes binarized weights and activations, while ReCU aims to address weights that are scarcely updated during BNN training. The incorporation of DeSign functions in both binarization methods produces positive outcomes, emphasizing the adaptable and versatile characteristics of the proposed activation.

\section{Conclusions}

The DeSign activation method introduces a design strategy to address information loss in Binary Neural Networks (BNNs). It improves BNN training accuracy without the additional computational overhead. DeSign offers two key advantages: it enhances BNNs by selectively capturing relevant information from binary convolutions, and it reduces the impact of real-valued learnable parameters in layers like batch-normalization, enabling the training of BNNs with full binary parameters and comparable performance. Simulations demonstrate that carefully selecting the spatial pattern significantly boosts BNN accuracy, improving baseline accuracy by up to 4.51$\%$. The proposed threshold kernel design methodology has the potential for further improvement using decision trees or end-to-end design approaches. Moreover, it can be extended to other real-valued activations beyond the ReLU.

\section{Acknowledgement}

This work was supported by the Vicerrectoría de Investigación Extensión of the Universidad Industrial de Santander, Colombia under the research project 3735.


%

\ifCLASSOPTIONcaptionsoff
  \newpage
\fi



%

{\small
\bibliographystyle{IEEEbib}
\bibliography{IEEEabrv, main}

\begin{thebibliography}{10}

\bibitem{szegedy2013deep}
Christian Szegedy, Alexander Toshev, and Dumitru Erhan,
\newblock ``Deep neural networks for object detection,''
\newblock {\em Advances in neural information processing systems}, vol. 26, 2013.

\bibitem{rawat2017deep}
Waseem Rawat and Zenghui Wang,
\newblock ``Deep convolutional neural networks for image classification: A comprehensive review,''
\newblock {\em Neural computation}, vol. 29, no. 9, pp. 2352--2449, 2017.

\bibitem{guo2018review}
Yanming Guo, Yu~Liu, Theodoros Georgiou, and Michael~S Lew,
\newblock ``A review of semantic segmentation using deep neural networks,''
\newblock {\em International journal of multimedia information retrieval}, vol. 7, no. 2, pp. 87--93, 2018.

\bibitem{mcdanel2017embedded}
Bradley McDanel, Surat Teerapittayanon, and H.T. Kung,
\newblock ``Embedded binarized neural networks,''
\newblock in {\em Proceedings of the 2017 International Conference on Embedded Wireless Systems and Networks}, USA, 2017, EWSN ’17, p. 168–173, Junction Publishing.

\bibitem{courbariaux2015binaryconnect}
Matthieu Courbariaux, Yoshua Bengio, and Jean-Pierre David,
\newblock ``Binaryconnect: Training deep neural networks with binary weights during propagations,''
\newblock {\em Advances in neural information processing systems}, vol. 28, 2015.

\bibitem{Kim2020BinaryDuo:}
Hyungjun Kim, Kyungsu Kim, Jinseok Kim, and Jae-Joon Kim,
\newblock ``Binaryduo: Reducing gradient mismatch in binary activation network by coupling binary activations,''
\newblock in {\em International Conference on Learning Representations}, 2020.

\bibitem{courbariaux2016binarized}
Matthieu Courbariaux, Itay Hubara, Daniel Soudry, Ran El-Yaniv, and Yoshua Bengio,
\newblock ``Binarized neural networks: Training deep neural networks with weights and activations constrained to+ 1 or-1,''
\newblock {\em arXiv preprint arXiv:1602.02830}, 2016.

\bibitem{xu2021recu}
Zihan Xu, Mingbao Lin, Jianzhuang Liu, Jie Chen, Ling Shao, Yue Gao, Yonghong Tian, and Rongrong Ji,
\newblock ``Recu: Reviving the dead weights in binary neural networks,''
\newblock in {\em Proceedings of the IEEE/CVF International Conference on Computer Vision}, 2021, pp. 5198--5208.

\bibitem{montesinos2022fundamentals}
Osval~Antonio Montesinos~L{\'o}pez, Abelardo Montesinos~L{\'o}pez, and Jose Crossa,
\newblock ``Fundamentals of artificial neural networks and deep learning,''
\newblock in {\em Multivariate Statistical Machine Learning Methods for Genomic Prediction}, pp. 379--425. Springer, 2022.

\bibitem{rastegari2016xnor}
Mohammad Rastegari, Vicente Ordonez, Joseph Redmon, and Ali Farhadi,
\newblock ``Xnor-net: Imagenet classification using binary convolutional neural networks,''
\newblock in {\em Computer Vision--ECCV 2016: 14th European Conference, Amsterdam, The Netherlands, October 11--14, 2016, Proceedings, Part IV}. Springer, 2016, pp. 525--542.

\bibitem{lin2020rotated}
Mingbao Lin, Rongrong Ji, Zihan Xu, Baochang Zhang, Yan Wang, Yongjian Wu, Feiyue Huang, and Chia-Wen Lin,
\newblock ``Rotated binary neural network,''
\newblock {\em Advances in neural information processing systems}, vol. 33, pp. 7474--7485, 2020.

\bibitem{cai2017deep}
Zhaowei Cai, Xiaodong He, Jian Sun, and Nuno Vasconcelos,
\newblock ``Deep learning with low precision by half-wave gaussian quantization,''
\newblock in {\em Proceedings of the IEEE conference on computer vision and pattern recognition}, 2017, pp. 5918--5926.

\bibitem{zhang2022dynamic}
Jiehua Zhang, Zhuo Su, Yanghe Feng, Xin Lu, Matti Pietik{\"a}inen, and Li~Liu,
\newblock ``Dynamic binary neural network by learning channel-wise thresholds,''
\newblock in {\em ICASSP 2022-2022 IEEE International Conference on Acoustics, Speech and Signal Processing (ICASSP)}. IEEE, 2022, pp. 1885--1889.

\bibitem{liu2020reactnet}
Zechun Liu, Zhiqiang Shen, Marios Savvides, and Kwang-Ting Cheng,
\newblock ``Reactnet: Towards precise binary neural network with generalized activation functions,''
\newblock in {\em European Conference on Computer Vision}. Springer, 2020, pp. 143--159.

\bibitem{schuchman1964dither}
Leonard Schuchman,
\newblock ``Dither signals and their effect on quantization noise,''
\newblock {\em IEEE Transactions on Communication Technology}, vol. 12, no. 4, pp. 162--165, 1964.

\bibitem{holesovsky2017compact}
Ondrej Holesovsky,
\newblock ``Compact convnets with ternary weights and binary activations,'' 2017.

\bibitem{ulichney1987digital}
Robert Ulichney,
\newblock {\em Digital halftoning},
\newblock MIT press, 1987.

\bibitem{6773262}
W.~M. Goodall,
\newblock ``Television by pulse code modulation,''
\newblock {\em The Bell System Technical Journal}, vol. 30, no. 1, pp. 33--49, 1951.

\bibitem{ulichney1999review}
Robert~A Ulichney,
\newblock ``Review of halftoning techniques,''
\newblock in {\em Color Imaging: Device-Independent Color, Color Hardcopy, and Graphic Arts V}. Spie, 1999, vol. 3963, pp. 378--391.

\bibitem{ulichney1999void}
Robert Ulichney,
\newblock ``The void-and-cluster method for dither array generation,''
\newblock {\em SPIE MILESTONE SERIES MS}, vol. 154, pp. 183--194, 1999.

\bibitem{ioffe2015batch}
Sergey Ioffe and Christian Szegedy,
\newblock ``Batch normalization: Accelerating deep network training by reducing internal covariate shift,''
\newblock in {\em International conference on machine learning}. pmlr, 2015, pp. 448--456.

\bibitem{qin2020binary}
Haotong Qin, Ruihao Gong, Xianglong Liu, Xiao Bai, Jingkuan Song, and Nicu Sebe,
\newblock ``Binary neural networks: A survey,''
\newblock {\em Pattern Recognition}, vol. 105, pp. 107281, 2020.

\bibitem{gelvez2020nonlocal}
Tatiana Gelvez and Henry Arguello,
\newblock ``Nonlocal low-rank abundance prior for compressive spectral image fusion,''
\newblock {\em IEEE Transactions on Geoscience and Remote Sensing}, vol. 59, no. 1, pp. 415--425, 2020.

\bibitem{bacca2021deep}
Jorge Bacca, Tatiana Gelvez-Barrera, and Henry Arguello,
\newblock ``Deep coded aperture design: An end-to-end approach for computational imaging tasks,''
\newblock {\em IEEE Transactions on Computational Imaging}, vol. 7, pp. 1148--1160, 2021.

\bibitem{ding2019regularizing}
Ruizhou Ding, Ting-Wu Chin, Zeye Liu, and Diana Marculescu,
\newblock ``Regularizing activation distribution for training binarized deep networks,''
\newblock in {\em Proceedings of the IEEE/CVF conference on computer vision and pattern recognition}, 2019, pp. 11408--11417.

\bibitem{qin2020forward}
Haotong Qin, Ruihao Gong, Xianglong Liu, Mingzhu Shen, Ziran Wei, Fengwei Yu, and Jingkuan Song,
\newblock ``Forward and backward information retention for accurate binary neural networks,''
\newblock in {\em Proceedings of the IEEE/CVF conference on computer vision and pattern recognition}, 2020, pp. 2250--2259.

\bibitem{zhou2016dorefa}
Shuchang Zhou, Yuxin Wu, Zekun Ni, Xinyu Zhou, He~Wen, and Yuheng Zou,
\newblock ``Dorefa-net: Training low bitwidth convolutional neural networks with low bitwidth gradients,''
\newblock {\em arXiv preprint arXiv:1606.06160}, 2016.

\end{thebibliography}
}

\end{document}